\documentclass[a4paper,twoside]{article}

\usepackage{epsfig}
 \graphicspath{{./figures/}}
 \DeclareGraphicsExtensions{.eps}
\usepackage{calc}
\usepackage{amssymb}
\usepackage{amstext}
\usepackage{amsmath}
\usepackage{amsthm}
\usepackage{multicol}
\usepackage{pslatex}
\usepackage{apalike}
\usepackage{SCITEPRESS} % Please add other packages that you may need BEFORE the SCITEPRESS.sty package.
\usepackage{multirow}
\usepackage{rotating}
 \usepackage{algpseudocode}
 \usepackage{listings}
 \usepackage{xcolor}
\begin{document}

\title{Ensemble UCT Needs High Exploitation}

\author{\authorname{S. Ali Mirsoleimani\sup{1,2}, Aske Plaat\sup{1} and Jaap van den Herik\sup{1}}
\affiliation{\sup{1}Leiden Centre of Data Science, Leiden University\\
 Niels Bohrweg 1, 2333 CA Leiden, The Netherlands}
\affiliation{\sup{2}Nikhef Theory Group, Nikhef\\
 Science Park 105, 1098 XG Amsterdam, The Netherlands}
%\email{\{f\_author, s\_author\}@ips.xyz.edu, t\_author@dc.mu.edu}
}


\abstract{Recent results have shown that the MCTS algorithm (a new, adaptive, randomized optimization algorithm) is effective in a remarkably diverse set of applications in Artificial Intelligence, Operations Research, and High Energy Physics. MCTS can find good solutions without domain dependent heuristics, using the UCT formula to balance exploitation and exploration. It has been suggested that the optimum in the exploitation-exploration balance differs for different search tree sizes: small search trees needs more exploitation; large search trees need more exploration. Small search trees occur in variations of MCTS, such as parallel and ensemble approaches. This paper investigates the possibility of improving the performance of Ensemble UCT by increasing the level of exploitation. As the search trees becomes smaller we achieve an improved performance. The results are important for improving the performance of large scale parallelism of MCTS.} 

\onecolumn \maketitle \normalsize \vfill

\section{\uppercase{Introduction}}
\label{sec:introduction}
\noindent Since its inception in 2006 \cite{Coulom2006}, the Monte Carlo Tree Search (MCTS) algorithm has gained much interest among optimization researchers. MCTS is a sampling algorithm that uses search results to guide itself through the search space, obviating the need for domain-dependent heuristics. Starting with the game of Go, an oriental board game, MCTS has achieved performance breakthroughs in domains ranging from planning and scheduling to high energy physics \cite{Chaslot2008,Kuipers2013,Ruijl2014}. The success of MCTS depends on the balance between exploitation (look in areas which appear to be promising) and exploration (look in areas that have not been well sampled yet). The most popular algorithm in the MCTS family which addresses this dilemma is the Upper Confidence Bound for Trees (UCT) \cite{Kocsis2006}.
	
\noindent As with most sampling algorithms, one way to improve the quality of the result is to increase the number of samples and thus enlarge the size of the MCTS tree. However, constructing a single large search tree with $t$ samples or playouts is a time consuming process. A solution for this problem is to create a group of $n$ smaller trees that each have $t/n$ playouts and search these in parallel. This approach is used in root parallelism \cite{Chaslot2008} and in Ensemble UCT \cite{fern2011ensemble}. In both, root parallelism and Ensemble UCT, multiple independent UCT instances are constructed. At the end of the search process, the statistics of all trees are combined to yield the final result \cite{Browne2012}. However, there is contradictory evidence on the success of Ensemble UCT \cite{Browne2012}. On the one hand, Chaslot et al. find that, for Go, Ensemble UCT (with $n$ trees of $t/n$ playouts each) outperforms a plain UCT (with $t$ playouts) \cite{Chaslot2008}. However, Fern and Lewis are not able to reproduce this result in other domains \cite{fern2011ensemble}, finding situations where a plain UCT outperforms Ensemble UCT given the same total number of playouts.
	
\noindent As already mentioned, the success of MCTS depends on the exploitation-exploration balance. Previous work by Kuipers et al. has argued that when the tree size is small, more exploitation should be chosen, and with larger tree sizes, high exploration is suitable \cite{Kuipers2013}. The main contribution of this paper is that we show that this idea can be used in Ensemble UCT to improve its performance.

The remainder of this paper is structured as follows: in section
\ref{sec:back} the required background information is briefly
discussed. Section \ref{sec:related} discusses related work. Section
\ref{sec:results} gives the experimental setup, together with the experimental results. Finally, a conclusion is given in Section \ref{sec:conclusion}. 

\section{\uppercase{Background}}
\label{sec:back}
\noindent Below we provide some background information on MCTS (Section \ref{sec:mcts}), Ensemble UCT (Section \ref{sec:ensembleuct}), and the game of Hex (Section \ref{sec:hex}).
\subsection{Monte Carlo Tree Search}
\label{sec:mcts}
\noindent The main building block of the MCTS algorithm is the search tree, where each
node of the tree represents a game position. The
algorithm constructs the search tree incrementally, expanding one node
in each iteration. Each iteration has four
steps~\cite{chaslot2008progressive}. (1) In the selection step, beginning
at the root of the tree, child nodes are selected successively
according to a selection criterion, until a leaf node is reached. (2)
In the expansion 
step, unless the selected leaf node ends the game, a random unexplored child of the leaf node is added to the
tree. (3) In the simulation step (also called playout step), the rest of the path to a final state is
completed by playing random moves. At the end a score $\Delta$ is
obtained that signifies the score of the chosen path through the state
space. (4) In the backprogagation step (also called backup step), the $\Delta$
value is propagated back through the traversed path in the tree, which updates the average
score (win rate) of a node. The number of times that each node in this
path is visited is incremented by one. Figure \ref{alg:mcts}
shows the general MCTS algorithm. In many MCTS implementations the UCT algorithm is chosen as the selection
criterion \cite{Kocsis2006}. 

\begin{figure}[t!]
\begin{algorithmic}
\Function{UCTSearch}{$r$,$m$}
\State $i\gets$ 1
\For{$i\leq m$}
 \State $n\gets$ select($r$)
 \State $n\gets$ expand($n$)
 \State $\Delta \gets$playout($n$)
 \State backup($n$,$\Delta$)
 \EndFor
\State \Return
\EndFunction
\end{algorithmic}
\caption{The general MCTS algorithm.}
\label{alg:mcts}
\end{figure}

\subsubsection{The UCT Algorithm}
\label{sec:uct}
\noindent The UCT algorithm provides a solution for the problem of exploitation
(look into promising areas) and exploration (look for promising
areas) in the selection phase of the MCTS algorithm~\cite{Kocsis2006}. A child
node $j$ is selected to maximize: 
\begin{equation}
UCT(j)=\overline{X}_{j}+C_{p}\sqrt{\frac{\ln(n)}{n_{j}}}
\end{equation}
where $\overline{X}_{j}=\frac{w_{j}}{n_{j}} $, $w_{j}$ is the number
of wins in child $j$, $n_{j}$ is the number of times child $j$ has
been visited, $n$ is the number of times the parent node has been
visited, and $C_{p}\geq0$ is a constant. The first term in UCT
equation is for exploitation and the second one is for
exploration. The level of exploration of the UCT
algorithm can be adjusted by the $C_{p}$ constant. (High $C_p$ means
more exploration.)

\subsubsection{Root Parallelism}
\begin{figure}[b!]
\centering
\includegraphics[width=.4\textwidth]{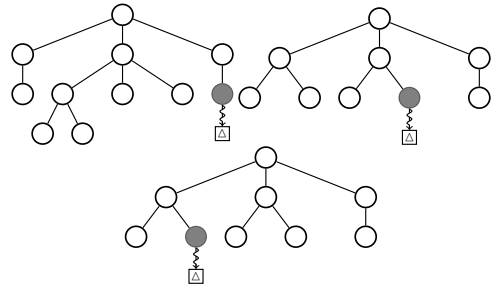}
\caption{Different independent UCT trees are used in root parallelism.}
\label{fig:root-parallel}
\end{figure}

\noindent Originally, root parallelism has been an UCT algorithm, viz. UCT in parallel. In root parallelism~\cite{Chaslot2008} each thread builds
simultaneously a private and independent MCTS search tree with a
unique random seed. When root parallelism wants to select the next
move to play, one of the threads collects the number of visits and
number of wins in the upper-most nodes of all trees and then computes the
total sum for each child~\cite{Chaslot2008}. Then, it selects a move
based on one of the possible policies. Figure \ref{fig:root-parallel} shows root parallelism. However, nowadays we have noted that UCT with root
parallelism is not algorithmically equivalent to plain UCT, but is
equivalent to Ensemble UCT~\cite{Browne2012}.

\subsection{Ensemble UCT}
\label{sec:ensembleuct}
\noindent Ensemble UCT is given its place in the overview article by \cite{Browne2012}. 
Table \ref{tab:ensmble} shows different possible configurations for
Ensemble UCT. Each configuration has its own benefits. The total number of playouts is $t$ and the size of ensemble (number of trees inside the ensemble) is $n$. It is supposed that $n$ processors are available which is equal to the ensemble size.
\begin{figure}[b!]
\begin{algorithmic}
\State $n \gets$ ensemble size or number of trees
\State $t \gets$ total number of playouts
\Function{EnsembleUCT}{$s$,$t$,$n$}
\State $m \gets t/n$
\State $i \gets 1$
\For{$i\leq n$}
	\State $r[i]\gets$ create an independent root node with state s
\EndFor
\State $i \gets 1$
\For{$i\leq n$}
	\State execute UCTSearch($r[i]$,$m$)
\EndFor
\State collect from all trees the number of wins and visits to the root's children. Then
compute the total sum of visits and wins for each child and store it to a new root $r^{\prime}$. 
\State \Return child with \textit{argmax} $w_{j}/n_{j}$ $j\in$ children of $r^{\prime}$
\EndFunction
\end{algorithmic}
\caption{The pseudo-code of Ensemble UCT.}
\label{alg:ensembleuct}
\end{figure}

\begin{table*}[t!]
\centering
\caption{Different possible configurations for Ensemble UCT. The ensemble size is $n$.}
\label{tab:ensmble}
\begin{tabular}{|c|l|l|c|l|c|}
\hline
\multicolumn{3}{|c|}{Number of playouts} & \multicolumn{2}{c|}{Playout speedup} & Strength speedup \\ \hline
\multicolumn{1}{|l|}{\multirow{2}{*}{UCT}} & \multicolumn{2}{l|}{Ensemble UCT} & \multicolumn{1}{l|}{\multirow{2}{*}{n cores}} & \multirow{2}{*}{1 core} & \multicolumn{1}{l|}{\multirow{2}{*}{}} \\ \cline{2-3}
\multicolumn{1}{|l|}{} & Each tree & Total & \multicolumn{1}{l|}{} & & \multicolumn{1}{l|}{} \\ \hline
$t$ & $t$ & $n\cdot t$ & 1 & \multicolumn{1}{c|}{$\frac{1}{n}$} & Yes \\ \hline
$t$ & $\frac{t}{n}$ &$t$ & $n$ & \multicolumn{1}{c|}{1} & ? \\ \hline
\end{tabular}
\end{table*}

\noindent The first line of the table shows the situation where Ensemble UCT has $n\cdot t$ playouts in total while UCT has only $t$ playouts. In this case, there would be no speedup in a parallel execution of the ensemble approach on $n$ cores, but the larger search effort would presumably result in a better search result. We call this use of
parallelism Strength speedup.

\noindent The second line of Table \ref{tab:ensmble} shows a different possible configuration for Ensemble UCT. In this case, the total number of playouts for
both UCT and Ensemble UCT is equal to $t$. Thus, each core searches a
smaller tree of size $\frac{t}{n}$. The search will be $n$ times
quicker (the ideal case). We call this use of parallelism \textit{Playout speedup}. It is important to note that in this configuration both approaches take the same amount of
time on a single core. However, there is still the question whether we can reach any
\textit{Strength speedup}. This question will be answered in this paper. Figure \ref{alg:ensembleuct} shows the pseudo-code of Ensemble UCT.

\subsection{The Game of Hex}
\label{sec:hex}
\noindent Hex is a game with a board of hexagonal cells
~\cite{Arneson2010}. Each player is represented by a color (White or
Black). Players take turns by placing a stone of their color on a cell
of the board. The goal for each player is to create a connected
chain of stones between the opposing sides of the board
marked by their colors. The first player to complete this path
wins the game.

\noindent In our implementation of the Hex game, a fast disjoint-set data structure
is used to determine the connected stones. Using this data structure
we have an efficient representation of the board position~\cite{Galil:1991:DSA:116873.116878}.

\section{\uppercase{Related Work}}
\label{sec:related}
% Marcolino and Matsubara~\cite{Marcolino2011} describe the simulation
% phase of UCT as a single agent playing against itself, and instead
% consider the effect of having multiple agents (i.e. multiple
% simulation policies). If the right subset of agents is chosen (or
% learned, as in ~\cite{Marcolino2011}), using multiple agents
% improves playing strength. Marcolino and Matsubara
% ~\cite{Marcolino2011} argue that the emergent properties of
% interactions between different agent types lead to increased
% exploration of the search space. However, finding the set of agents
% with the correct properties (i.e. those that increase playing
% strength) is computationally intensive.

% I do not think this paper is related to our work

\noindent Chaslot et al. provided evidence that, for Go,
root parallelism with $n$ instances of $\frac{t}{n}$ iterations each outperforms
plain UCT with $t$ iterations, i.e., root parallelism (being a form of Ensemble UCT) outperforms
plain UCT given the same total number of iterations~\cite{Chaslot2008}. However, in
other domains, Fern and
Lewis do not find this result \cite{fern2011ensemble}. 

\noindent \cite{Soejima2010} also analyzed the performance of root
parallelism in detail. They found
that a majority voting scheme gives better performance than the
conventional approach of playing the move with the greatest total
number of visits across all trees. They suggested that
the findings in \cite{Chaslot2008} are explained by the fact that
root parallelism performs a shallower search, making it easier for UCT to escape from local
optima than the deeper search performed by plain UCT. In root
parallelism each process does not build a search tree larger than
sequential UCT; each process has a
local tree that contains characteristics that are different from the
others. 

\noindent Fern and Lewis‍ thoroughly investigated an Ensemble UCT
approach in which multiple instances of UCT were run independently and
their root statistics were combined to yield the final result~\cite{fern2011ensemble}. 
So, our task is to explain the differences in these work.

\begin{table*}[ht!]
\centering
\caption{The performance evaluation of Ensemble UCT vs. plain UCT based on win rate.}
\label{tab:performance}
\begin{tabular}{|c|c|c|c|}
\hline
Approach & Win (\%) & \begin{tabular}[c]{@{}c@{}}Performance vs. \\ plain UCT\end{tabular} & \begin{tabular}[c]{@{}c@{}}Strength \\ Speedup\end{tabular} \\ \hline
\multirow{3}{*}{Ensemble UCT} & $<50$ & Worse than & No \\ \cline{2-4} 
 & $=50$ & As good as & No \\ \cline{2-4} 
 & $>50$ & Better than & Yes \\ \hline
\end{tabular}
\end{table*}

\section{\uppercase{Empirical Study}}
\label{sec:results}
In this section, the experimental setup is described and then
the experimental results are presented. 
\subsection{Experimental Setup}
The Hex board is represented by a disjoint-set. This data structure has
three operations \textit{MakeSet}, \textit{Find} and
\textit{Union}. In the best case, the amortized time per operation is
$O\left ( \alpha \left ( n \right ) \right )$. The value of $\alpha
\left ( n \right )$ is less than 5 for all remotely practical values
of $n$ \cite{Galil:1991:DSA:116873.116878}. 

In Ensemble UCT, each tree performs a completely independent UCT
search with a different random seed. To determine the next move to
play, the number of wins and visits of the root's children of all trees are
collected and the total sum of wins and visits for each child is computed. The
child with the largest number of wins/visits is selected. 

The plain UCT algorithm and Ensemble UCT are implemented in C++. In order to make our experiments
as realistic as possible, we use a custom developed game playing
program for the game of Hex \cite{Mirsoleimani2014b,Mirsoleimani2015a}. This
program is highly optimized, and reaches a speed of more than 40,000 playouts per
second per core on a 2,4 GHz Intel Xeon processor. The source code of the
program is available online.
%\footnote{Source code is available at https://github.com/AAA/}
\footnote{Source code is available at https://github.com/mirsoleimani/paralleluct/} 

As Hex is a 2-player game, the playing strength of Ensemble UCT is
measured by playing versus a plain UCT with the same number of
playouts. We expect to see an improvement for Ensemble
UCT playing strength against plain UCT by choosing 0.1 as the value of $C_{p}$ (high exploitation) when the number of playouts is small. 
The value of $C_{p}$ is set to 1.0 for plain UCT (high exploration). Note that for the
purpose of this research it
is not important to find the optimal value of $C_{p}$, but just to 
to show the difference in effect on the performance. 

Our experimental results show the percentage of wins for Ensemble UCT with a
particular ensemble size and a particular $C_{p}$ value against plain UCT. Each data point show the average of 200 games with 
the corresponding 99\% confidence interval. Table
\ref{tab:performance} summarizes how the performance of Ensemble
UCT versus plain UCT is evaluated. The concept of \textit{high exploitation for small UCT tree} is significant if Ensemble UCT reaches more than 50\% win rate. (The next section will shown that this is indeed the case.)

The board size for Hex is 11x11. In our experiments the maximum
ensemble size is $2^{8}=256$. Thus, for $2^{17}$ playouts, when
the ensemble size is 1 there are $2^{17}$ playouts per tree and when the
ensemble size is $2^{6}=64$ the number of playouts per tree is
$2^{11}$. Throughout the experiments the ensemble size is multiplied by
a factor of two. 

The results were measured on a dual socket machine with 2 Intel {\em
 Xeon\/} E5-2596v2 processors running at 2.40GHz. Each processor has
12 cores, 24 hyperthreads and 30 MB L3 cache. Each physical core has
256KB L2 cache. The pack TurboBoost frequency is 3.2 GHz. The machine
has 192GB physical memory. Intel's {\em icc 14.0.1} compiler is used
to compile the program.

\subsection{Experimental Results}
Below we provide our experimental results. We distinguish them into hidden exploration in Ensemble UCT (\ref{sec:hidden}) and exploitation-exploration trade-off for Ensemble UCT (\ref{sec:trade-off}).
\begin{figure*}
\begin{minipage}[b]{0.5\linewidth}
%\begin{figure}
% \vspace{-0.2cm}
 \centering
\includegraphics[width=7cm]{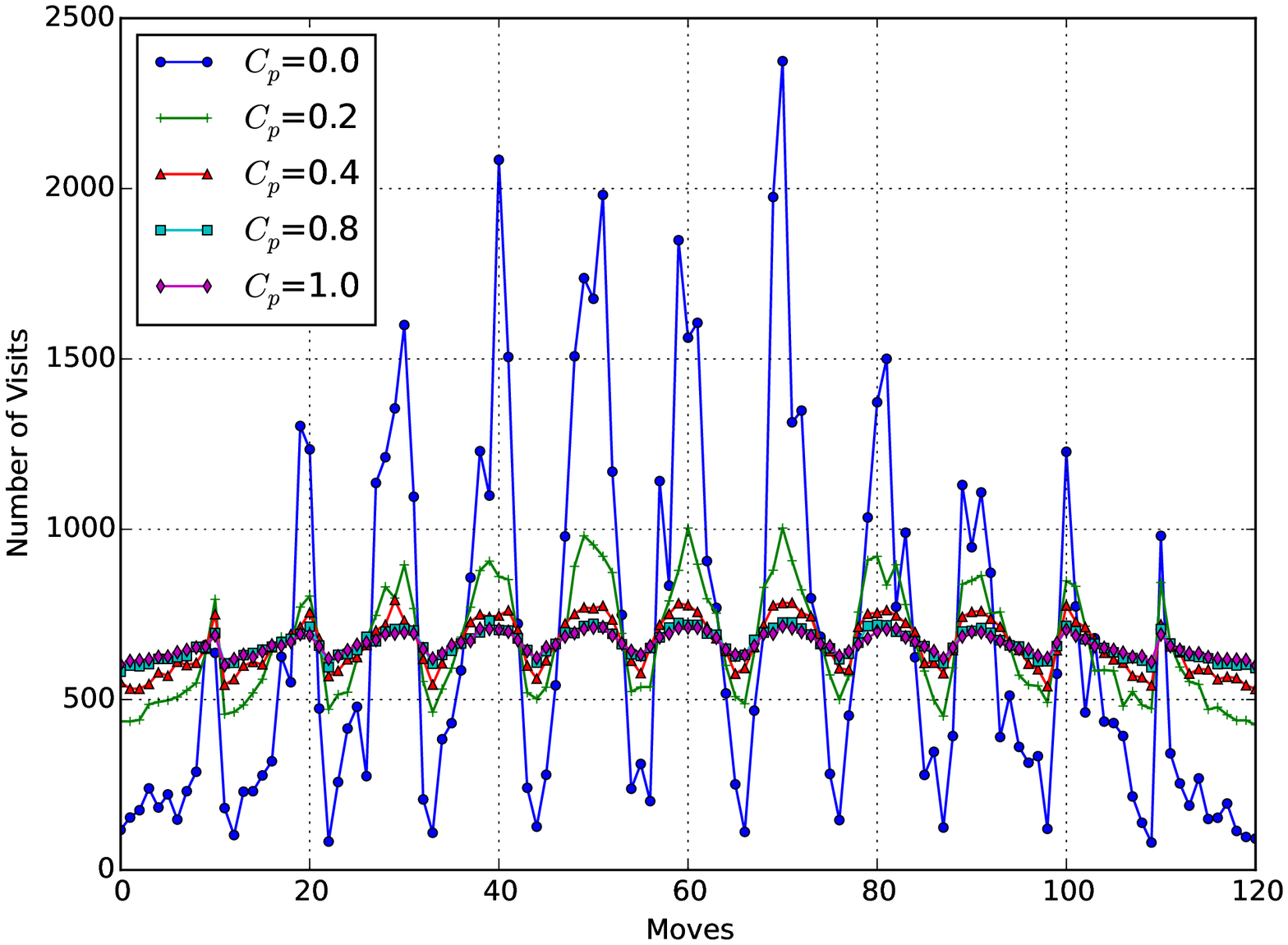}
 \caption{The number of visits for root's children in Ensemble UCT. Each child represents an available move on the Hex board. The total number of playouts is 80,000. There are 8 trees in the ensemble. Each thread of them has 10,000 playouts.}
 \label{fig:icaart16-1}
 \vspace{-0.4cm}
%\end{figure}
\end{minipage}
\qquad
\begin{minipage}[b]{0.5\linewidth}
%\begin{figure}
% \vspace{-0.2cm}
 \centering
\includegraphics[width=7cm]{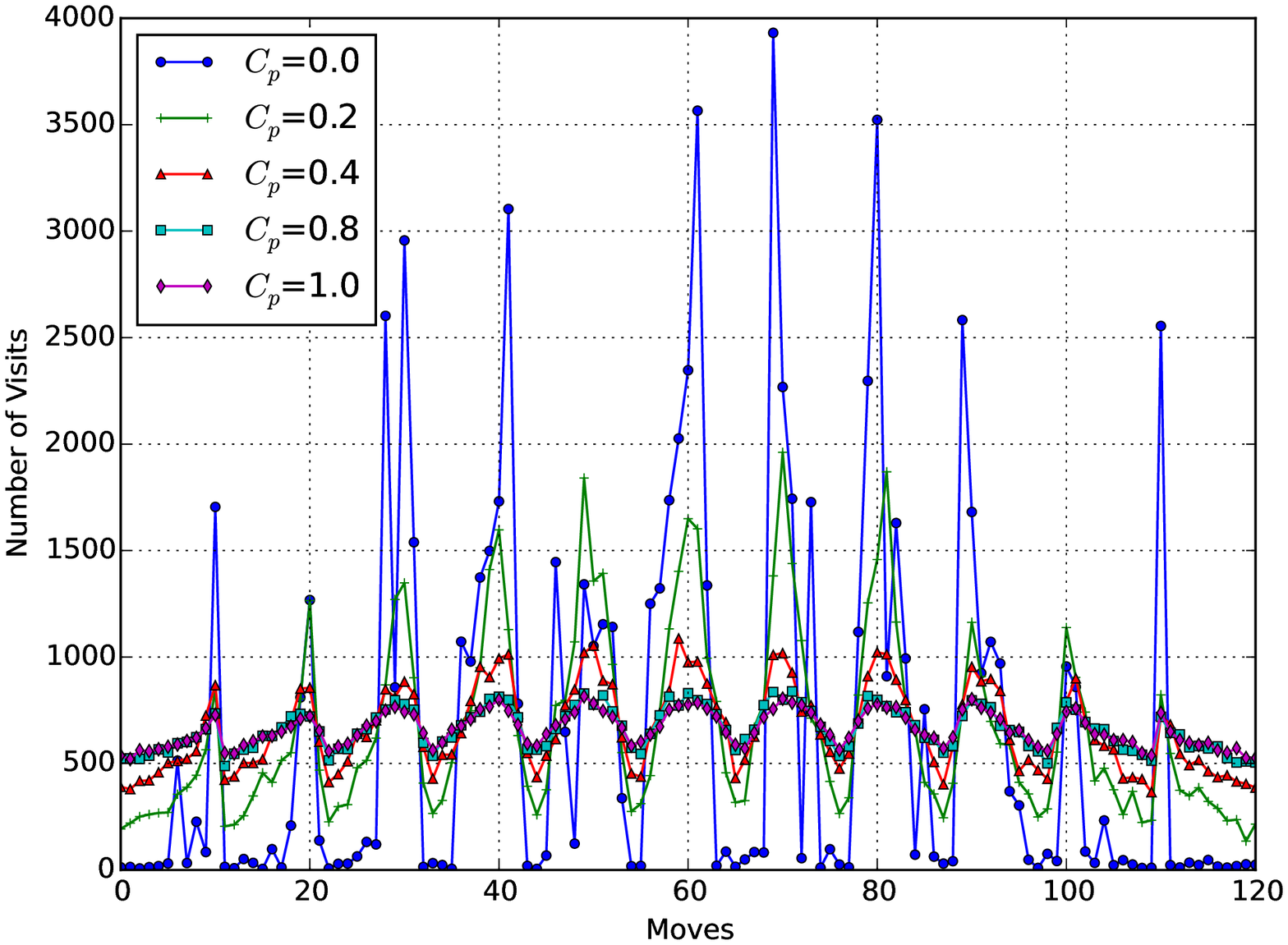}
 \caption{The number of visits for root's children in plain UCT. Each child represents an available move on the Hex board. The total number of playouts is 80,000.}
 \label{fig:icaart16-2}
 \vspace{-0.1cm}
%\end{figure}
\end{minipage}
\end{figure*}
\subsubsection{Hidden Exploration in Ensemble UCT}
\label{sec:hidden}
It is important to understand that Ensemble UCT has a hidden exploration
factor by nature. The reasons are (1) each tree in Ensemble UCT is
independent, and (2) an ensemble of trees contains more exploration than
a single UCT search of comparable size would have. Figure
\ref{fig:icaart16-1} and \ref{fig:icaart16-2} show how
this hidden exploration factor affects the performance of both, Ensemble UCT and plain UCT, in the decision making process at the level of individual moves. 
%The results are measured for different $C_{p}$ values. In this experiment the maximum number of playouts is 80000. 
In the graphs, a board position of game of Hex is used where moves around number 60 are
the best ones. Therefore, the graph of the number of visits for children
of the root with a bell shape around move number 60 is the desirable
one. 

In Figure \ref{fig:icaart16-1} the graph shows the number of visits
for the root's children when using Ensemble UCT. The number of trees
in the ensemble is 8 and the number of playouts per tree is 10,000. In
Figure \ref{fig:icaart16-2} the performance of palin UCT with 80,000
playouts is shown. Both algorithms have the same total number of
playouts. However, for $C_{p}=0$, which means the exploration part of 
UCT formula is turned off, comparing two graphs show that all root's
children in the Ensemble UCT have received at least a few visits while in UCT there are many unvisited children. The difference between
performance of these two approaches for other $C_{p}$ values is also
shown. This experiment shows the hidden exploration in Ensemble
UCT: in Figure \ref{fig:icaart16-1} the $C_{p}=0$ exploitation peaks reach up to 2400 visits, while in Figure \ref{fig:icaart16-2} the
$C_{p}=0$ exploitation peaks reach up to 4000 visits. Ensemble UCT dulls
the exploitation peaks.

\subsubsection{Exploitation-Exploration trade-off for Ensemble UCT}
\label{sec:trade-off}
\begin{figure*}[!t]
\begin{minipage}[b]{0.5\linewidth}
%\begin{figure}
 \vspace{-0.2cm}
 \centering
\includegraphics[width=7cm]{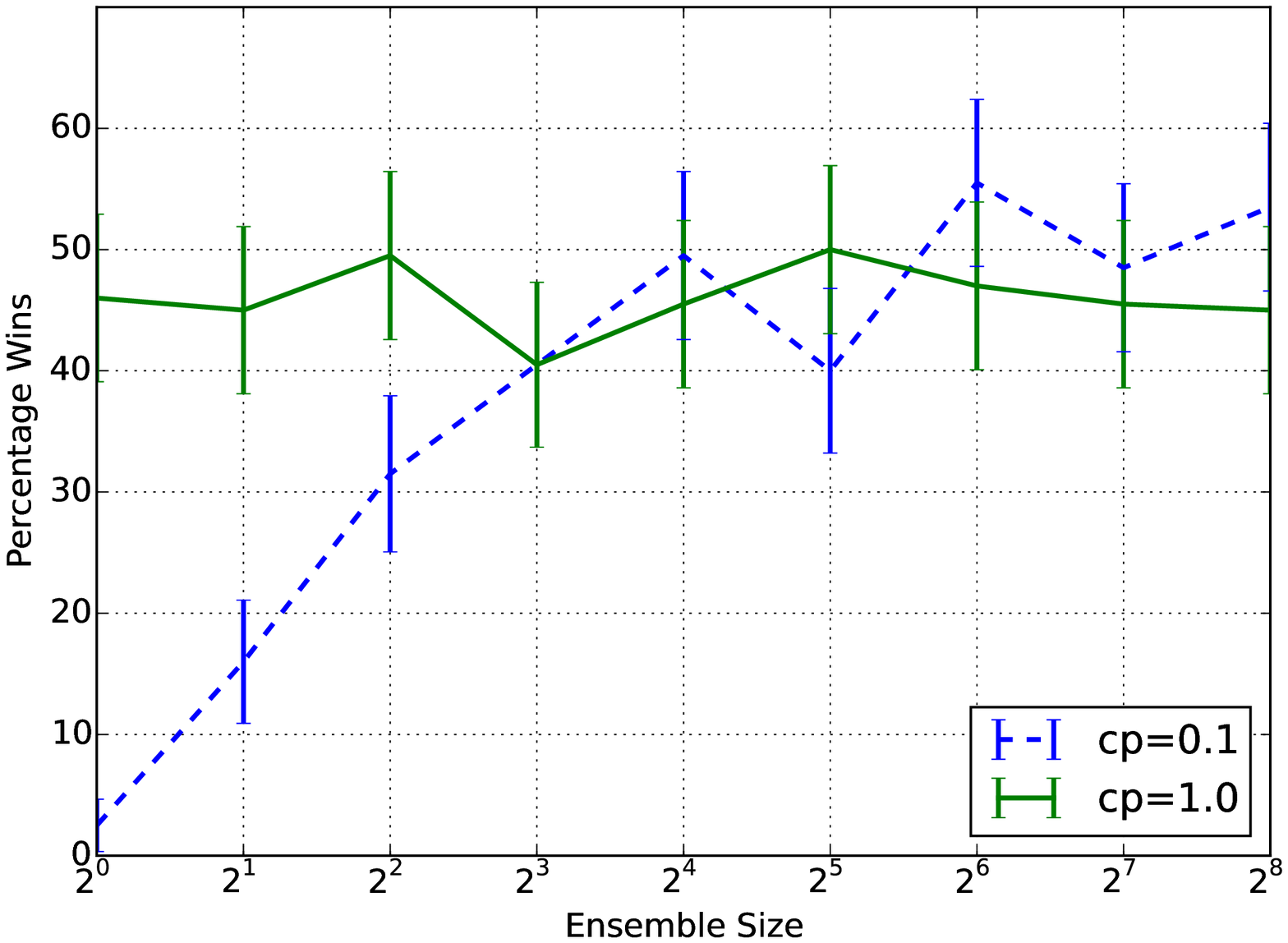}
\caption{The total number of playouts for both plain UCT and ensemble UCT is $2^{17}=131072$. The percentage of wins for ensemble UCT is reported. The value of $C_{p}$ for plain UCT is always 1.0 when playing against Ensemble UCT. To the left few large UCT trees, to the right many small UCT trees.}
\label{fig:icaart16-3}
 \vspace{-0.1cm}
%\end{figure}
\end{minipage}
\qquad
\begin{minipage}[b]{0.5\linewidth}
%\begin{figure}
 \vspace{-0.2cm}
 \centering
\includegraphics[width=7cm]{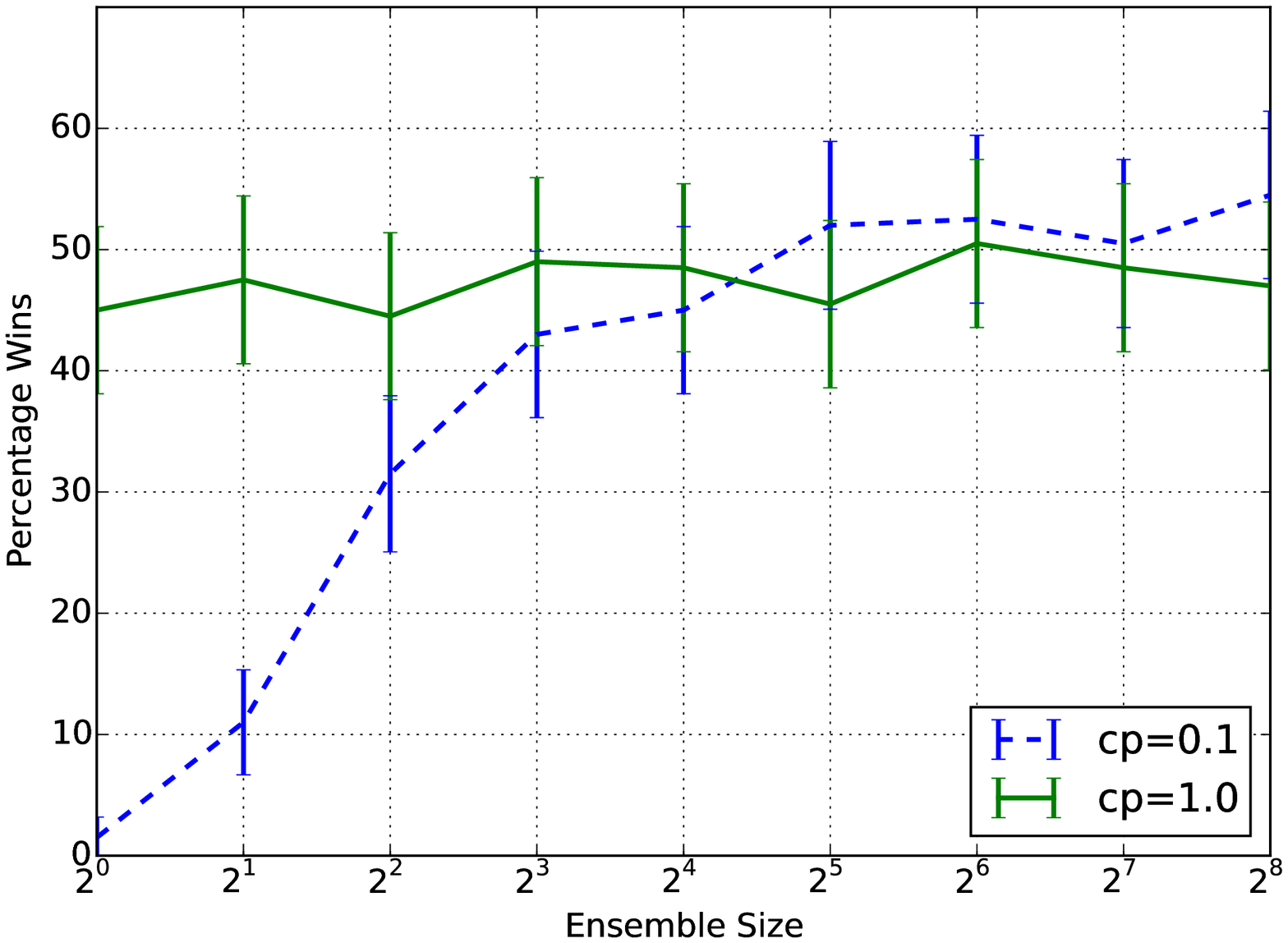}
\caption{The total number of playouts for both plain UCT and ensemble UCT is $2^{18}=262144$. The percentage of wins for ensemble UCT is reported. The value of $C_{p}$ for plain UCT is always 1.0 when playing against Ensemble UCT. To the left few large UCT trees, to the right many small UCT trees.}
\label{fig:icaart16-4}
 \vspace{-0.1cm}
%\end{figure}
\end{minipage}
\end{figure*}

In Figures \ref{fig:icaart16-3} and \ref{fig:icaart16-4}, from the left side to the right side of a graph, the
ensemble size (number of search trees per ensemble) increases by a
factor of two and the number of playouts per tree (tree size)
decreases by the same factor. Thus, at the most right hand
side of the graph we have the largest ensemble with smallest
trees. The total number of playouts always remains the same throughout an
experiment for both Ensemble UCT and plain UCT. The value of $C_{p}$
for plain UCT is always 1.0 which means high exploration. 

Figure~\ref{fig:icaart16-3} shows the \textit{relations} between the value of
$C_{p}$ and the ensemble size, when both plain UCT and Ensemble UCT
have the same number of total playouts. Moreover, Figure \ref{fig:icaart16-3} shows
 the \textit{performance} of Ensemble UCT for different values of
$C_{p}$. It shows that when $C_{p}=1.0$ (highly
explorative) Ensemble UCT performs as good as or mostly worse than
 plain UCT. When Ensemble UCT uses $C_{p}=0.1$ (highly exploitative) then for small
ensemble sizes (large sub-trees) the performance of Ensemble UCT sharply drops down. By
increasing the ensemble size (smaller sub-trees), the performance of Ensemble UCT keeps
growing improving until it becomes as good as or even better than
 plain UCT. 

In order to investigate the effect of enlarging number of playouts on
the performance of Ensemble UCT, second experiments is conducted
using $2^{18}$ playouts. Figure \ref{fig:icaart16-4} shows that 
when for this large number of playouts the value of $C_{p}=1.0$ is high
(i.e., highly explorative) the performance of Ensemble UCT can not be better that
 plain UCT. While for a small value of $C_{p}=0.1$ (i.e., highly exploitative) the performance of
Ensemble UCT is almost always better than plain UCT after ensemble
size is $2^{5}$.

\section{\uppercase{Conclusion}}
\label{sec:conclusion}
This paper describes an empirical study on Ensemble UCT
with different sets of configurations for ensemble size, tree size and
exploitation-exploration trade-off. Previous studies on
Ensemble UCT/root parallelism provided inconclusive evidence on the effectiveness of Ensemble UCT \cite{Chaslot2008,fern2011ensemble,Browne2012}. Our results suggest that the reason lies in the
exploration-exploitaiton trade-off in relation to the \textit{size of the
sub-trees}. Our results provide clear
evidence that the performance of Ensemble UCT is improved by selecting
higher exploitation for smaller search trees 
given a fixed time bound or number of simulations.

This work is motivated, in part, by the observation in \cite{Chaslot2008} of super-linear speedup in root parallelism. Finding super-linear speedup in two-agent games occurs infrequently. Most studies in parallel game-tree search report a battle against search overhead, communication overhead, and synchronization overhead (see, e.g., \cite{RomeinPhDthesis2001}. For super-linear speedup to occur, the parallel search must search {\em fewer} nodes than the sequential search. In most algorithms, parallellizations suffer because parts of the tree are searched with less information than is available in the sequential search, causing {\em more} nodes to be expanded. This study has shown how the situation that the parallel search tree is smaller than the sequential search tree can indeed occur in MCTS. The ensemble of the independent (parallel) sub-trees can be smaller than the monolithic total tree. When $C_p$ is chosen low (i.e., exploitative) the Ensemble search runs efficiently, where the monolithic plain UCT search is less efficient (see Figures 6 and 7).

For future work, we will explore other parts of the parameter space, to find optimal $C_{p}$ settings for different combinations of tree size and ensemble size. Also we will study the effect in different domains. Even more important is the effect of $C_{p}$ in tree parallelism \cite{Chaslot2008}.

\section*{\uppercase{Acknowledgements}}
This work is supported in part by the ERC Advanced Grant no. 320651, ``HEPGAME.''
%This work is supported in part by the AAA no. BBB, ``CCC.''

%
% ---- Bibliography ----
%
% the back matter
%\vfill
\bibliographystyle{apalike}
{\small
\bibliography{Bib-icaart16}}

\end{document}